\documentclass[conference]{IEEEtran}
\IEEEoverridecommandlockouts

\usepackage{cite}
\usepackage{amsmath,amssymb,amsfonts}
\usepackage{algorithmic}
\usepackage{graphicx}
\usepackage{textcomp}
\usepackage{xcolor}
\usepackage{amsthm}
\usepackage{comment}

\def\BibTeX{{\rm B\kern-.05em{\sc i\kern-.025em b}\kern-.08em
    T\kern-.1667em\lower.7ex\hbox{E}\kern-.125emX}}
\makeatletter
\newcommand{\linebreakand}{%
  \end{@IEEEauthorhalign}
  \hfill\mbox{}\par
  \mbox{}\hfill\begin{@IEEEauthorhalign}
}
\makeatother

\theoremstyle{definition}

\begin{document}

\title{DeviceScope: An Interactive App to Detect and Localize Appliance Patterns in Electricity Consumption Time Series}

\author{\IEEEauthorblockN{Adrien Petralia}
\IEEEauthorblockA{\textit{EDF R\&D - Université Paris Cité} \\
adrien.petralia@gmail.com}
\and
\IEEEauthorblockN{Paul Boniol}
\IEEEauthorblockA{\textit{Inria, ENS, PSL, CNRS} \\
paul.boniol@inria.fr}
\and
\IEEEauthorblockN{Philippe Charpentier}
\IEEEauthorblockA{\textit{EDF R\&D} \\
philippe.charpentier@edf.fr}
\and
\IEEEauthorblockN{Themis Palpanas}
\IEEEauthorblockA{\textit{Université Paris Cité - IUF} \\
themis@mi.parisdescartes.fr}
}

\maketitle

\begin{abstract}
In recent years, electricity suppliers have installed millions of smart meters worldwide to improve the management of the smart grid system.
These meters collect a large amount of electrical consumption data to produce valuable information to help consumers reduce their electricity footprint.
However, having non-expert users (e.g., consumers or sales advisors) understand these data and derive usage patterns for different appliances has become a significant challenge for electricity suppliers because these data record the aggregated behavior of all appliances.
At the same time, ground-truth labels (which could train appliance detection and localization models) are expensive to collect and extremely scarce in practice.
This paper introduces DeviceScope~\cite{DeviceScope_app}, an interactive tool designed to facilitate understanding smart meter data by detecting and localizing individual appliance patterns within a given time period.
Our system is based on CamAL (Class Activation Map-based Appliance Localization), a novel weakly supervised approach for appliance localization that only requires the knowledge of the existence of an appliance in a household to be trained. 
This paper appeared in ICDE 2025.
\end{abstract}

\maketitle

\section{Introduction}
\label{sec:intro}

In the last decade, millions of smart meters have been installed in households across the globe by electricity providers. 
These meters capture over time the total electricity consumed in a household. 
The collected data, though, correspond to the aggregated signatures of the appliances that are simultaneously active in the household, making it hard to understand and distinguish the individual appliances.
Thus, tools that enable consumers and suppliers to understand and analyze these data more easily are needed.

One way of understanding these data is to recognize if and when an appliance has been used by looking only at the aggregated signal.
This challenge, called Appliance Detection~\cite{eEnergy_ApplDetection}, is part of Non-Intrusive Load Monitoring (NILM), a well-known and growing research field that aims to identify the power consumption, pattern, or on/off state activation of individual appliances using only the total recorded consumption signal~\cite{themis_reviewnilm}.
Detecting an appliance can be seen as a step of NILM-based methods~\cite{huquet_activation}, and numerous approaches have been proposed in the literature~\cite{themis_reviewnilm}. 
Nevertheless, these solutions must be trained using individual appliance power, i.e., knowing the exact state of activation and consumption power of each appliance for each timestamp (e.g., strong label). 
Yet, gathering such data is expensive, as each appliance needs to be monitored with sensors to measure its individual consumption. 
In practice, the available information is merely the activation (or not) of an appliance within a time frame, and most of the time, only the possession of the appliance by a household (e.g., weak label) is available. 
Note that NILM approaches cannot operate with such scarce labels: trying to train a NILM solution with only one label for the entire series (e.g., by replicating the label for all time steps) implies that it can no longer be used to localize an appliance; indeed, NILM solutions provide a probability of detection for each individual timestamp to be able to localize it.

\begin{figure}
    \centering
    \includegraphics[width=0.98\linewidth]{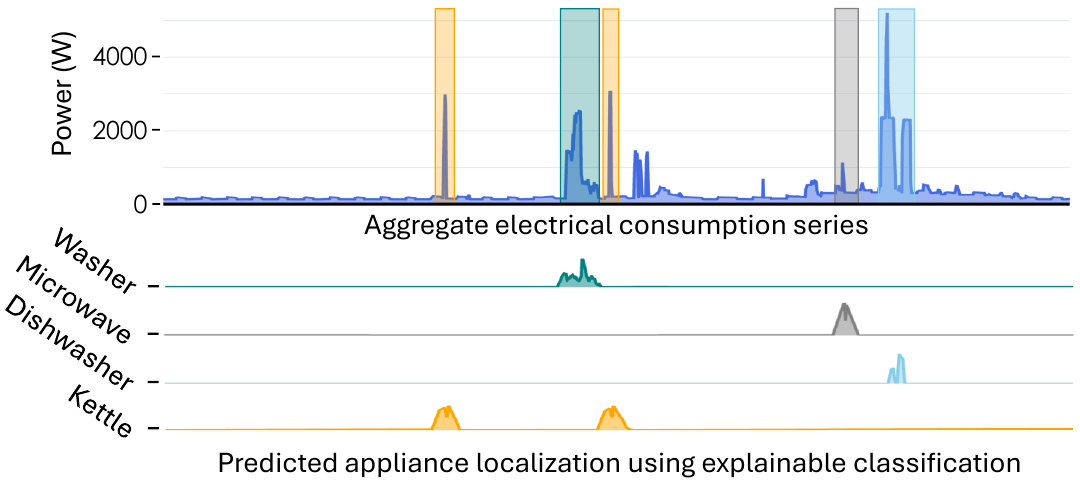}
    \caption{Illustration of the localization of different appliances in an aggregate consumption series. 
    }
    \vspace{-0.4cm}
    \label{fig:introfig}
\end{figure}

Some works have studied the problem of detecting the presence of appliances in smart meter data using weakly supervised approaches~\cite{eEnergy_ApplDetection, VLDB_TransApp}, where the appliance detection problem is cast as a time series classification problem~\cite{tsc_bakeoffredux_2024}: the classifier is trained using only one label for an electrical time series, i.e., by only knowing whether the appliance has been switched ON in a given time period, or by using the information on the presence of the appliance in the household.
Although these methods show promising results in detecting \emph{if} an appliance has been used, they cannot determine \emph{when} the device has been switched on.
Recent studies have shown that classification-based explainability methods can be used to understand a classifier's decision-making process by identifying the part of a time series that contributed to the label prediction.
These approaches have been tested on time series data~\cite{Wang2016TimeSC, dcam}, but have never been applied for appliance localization.

In this work, we introduce DeviceScope~\cite{DeviceScope_app}, 
a system that helps non-expert users understand their electrical consumption data.
DeviceScope is built upon CamAL (Class Activation Map-based Appliance Localization), a novel weakly supervised method for appliance pattern localization that requires only the appliance's possession label for training.
Our system enables users to (i) browse recorded electricity consumption series, (ii) as depicted in Figure~\ref{fig:introfig}, detect and localize different common appliances in a given period of time, and (iii), compare the performance of CamAL against other methods.


Overall, this demonstration has two goals: (i) facilitate the visualization and understanding of smart meter consumption data through appliance detection and localization methods, (ii) showcase the performance of CamAL and, for the first time, (iii) introduce the use of explainability-based approaches for solving the NILM problem. 

\section{Proposed Approach: CamAL} 
\label{sec:pipeline}

\begin{figure}
    \centering
    \includegraphics[width=1\linewidth]{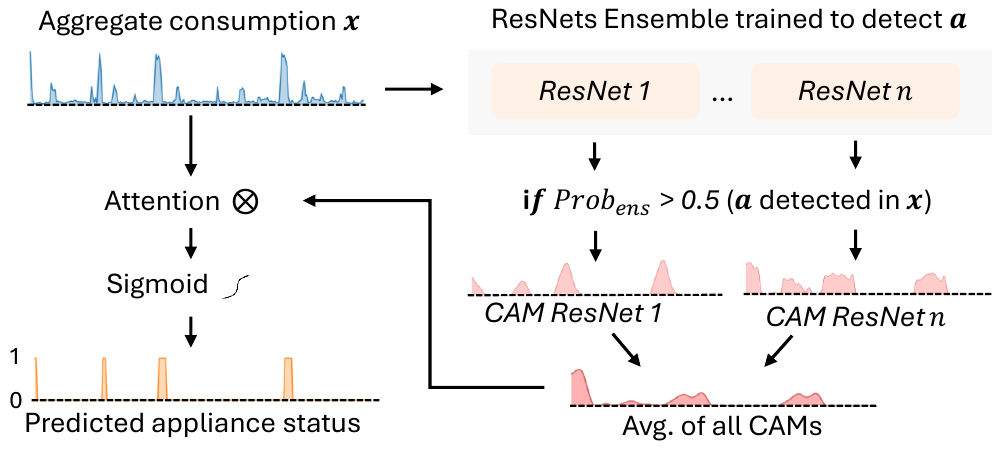}
    \vspace{-0.5cm}
    \caption{Illustration of the proposed CamAL approach.}
    \vspace{-0.5cm}
    \label{fig:pipeline}
\end{figure}

We now describe CamAL, the core of our system, that enables the detection and localization of appliance patterns in aggregated consumption series.
CamAL is composed of two parts (see Figure~\ref{fig:pipeline}): (1) an ensemble of deep-learning classifiers that performs the detection and (2) an explainability-based module that localizes the appliance (when detected).

\subsection{Step 1: Appliance Detection}
\label{subsec:appldetection}

Detecting if an appliance has been used in a period of time can be cast as a time series classification (TSC) problem~\cite{eEnergy_ApplDetection}.
To do so, a classifier is trained in a binary supervised manner to detect the presence of an appliance using only one label (0 or 1) for an entire series.
Based on previous studies~\cite{eEnergy_ApplDetection, VLDB_TransApp}, deep learning classifiers (specifically convolutional-based) are the most efficient and accurate solutions for tackling this task.
Thus, our system is based on an ensemble of convolutional residual networks (ResNets) to detect if an appliance pattern is present in a consumption series.

\noindent{\bf[ResNets Ensemble]} The original Residual Network architectures introduced for TSC~\cite{Wang2016TimeSC} consist of stacked residual blocks with residual connections, where each block contains 1D convolutional layers with the same kernel sizes.
At the network's end, a global average pooling (GAP) is followed by a linear layer to perform classification.
We leverage this baseline to an ensemble of 5 networks differing in kernel sizes within the convolutional layers. 
Specifically, we trained multiple networks with kernel sizes $k$ where $k \in {5, 7, 9, 15}$. We then selected the networks that best detected specific appliances. 
This approach is based on the premise that varying kernel sizes change the receptive fields of the convolutional neural network (CNN), offering different levels of explainability.

\noindent{\textbf{[Training Phase]}} 
First, we resample the datasets to a common frequency (1min).
We process the data by dividing each household's electricity consumption into subsequences and omitting subsequences with missing data.
For the IDEAL dataset, we assign to each subsequence the label of possession of the appliance provided in the survey questionnaire.
For the two other datasets (UKDALE and REFIT), we use the corresponding disaggregated appliance load curve to assign to each sub-sequence a positive or negative label, i.e., appliance presence or not: only this label is used for training (no information about the appliance's electricity consumption).
In addition, we note that we carefully use distinct houses for training and testing our solution and that the provided time series available for demonstrating our system are from distinct houses than the one used for training our solution.


\begin{figure}[tb]
    \centering
    \includegraphics[width=\linewidth]{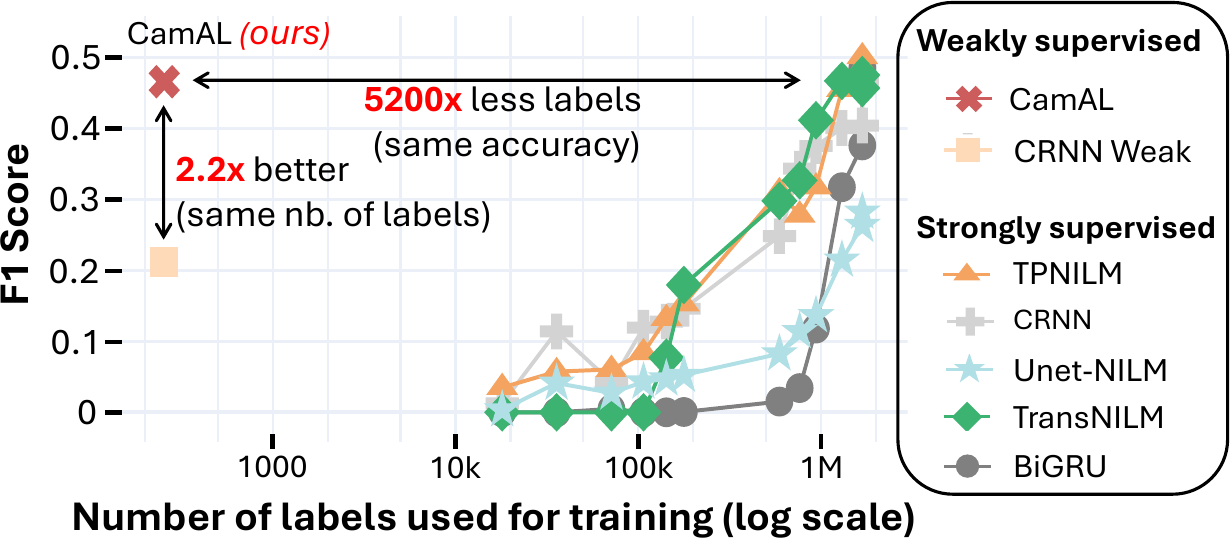}
    \vspace{-0.3cm}
    \caption{CamAL's localization accuracy vs. number of labels used for training compared to other baselines (Dishwasher case of the IDEAL dataset).}
    \vspace{-0.3cm}
    \label{fig:expresults}
\end{figure}

\subsection{Step 2: Appliance Pattern Localization}
\label{subsec:explainability}

Identifying the discriminative features that influence a classifier's decision-making process has been extensively studied. 
In deep-learning classification, different methods have been proposed to highlight (i.e., localize) the parts of an input instance that contribute the most to the final decision of the classifier~\cite{cam, gradcam, dcam}.
Based on this previous work, we developed CamAL, a method specifically designed to localize appliance patterns in electricity consumption series.

\begin{enumerate}
    \item \textbf{Ensemble Prediction}: The input sequence $\boldsymbol{x}$ is processed by an ensemble of ResNet models. 
    Each model independently estimates the probability that the target appliance is detected within the given sequence. 
    The final ensemble prediction is obtained by averaging the individual model probabilities:
    $\text{Prob}_{\text{ens}} = \frac{1}{N} \sum_{n=0}^N \text{Prob}_n$,
    where $N$ is the number of models in the ensemble, and $\text{Prob}_n$ is the prediction from the $n$-th model.

    \item \textbf{Appliance Detection}: If the ensemble probability exceeds a threshold (e.g., $\text{Prob}_{\text{ens}} > 0.5$), the appliance is considered detected in the current window.

    \item \textbf{CAM Extraction}: For each model $n$, we extract the Class Activation Map (CAM) corresponding to class 1 (i.e., appliance detected).
    In the case of a univariate time series, the CAM for class $c$ at timestamp $t$ is given by:
    $\text{CAM}_c(t) = \sum_{k} w_k^c \cdot f_k(t)$,
    where $w_k^c$ are the weights associated with the $k$-th filter for class $c$, and $f_k(t)$ is the activation of the $k$-th feature map at time $t$.

    \item \textbf{CAM processing}: Each extracted CAM, $\text{CAM}_n$, is first normalized to the range $[0,1]$, resulting in $\widetilde{\text{CAM}}_n(t)$. 
    The final output is then given as the average of the normalized CAMs across all models:
    $\text{CAM}_{\text{avg}}(t) = \frac{1}{N} \sum_{n=1}^{N} \widetilde{\text{CAM}}_n(t)$ where  $N$ is the total number of models in the ensemble.

    \item \textbf{Attention Mechanism}: $\text{CAM}_{avg}$ serves as an attention mask, highlighting the ensemble decision for each timestamp. 
    We apply this mask to the input sequence through point-wise multiplication and pass the results through a sigmoid activation function to map the values in $[0,1]$: 
    $\mathbf{s}(t) = \text{Sigmoid}(\text{CAM}_{\text{avg}}(t) \circ \mathbf{x}(t))$.

    \item \textbf{Appliance Status}: The obtained signal is then rounded to obtain binary labels ($1$ if $s(t) \geq 0.5$), 
    resulting in a binary time series $\hat{y}(t)$ that indicates the predicted status of the appliance for each timestep.
\end{enumerate}

\subsection{CamAL Results}
\label{subsec:results}

We compared our method against the SotA methods for NILM in terms of accuracy and number of labels needed for training (see Figure~\ref{fig:expresults}), including 5 NILM seq2seq approaches (which require one label by timestep) and one weakly supervised approach. 
The results show that our method is 2.2x better regarding F1-Score accuracy than the only other weakly supervised baseline.
In addition, to achieve the same performance as CamAL, NILM-based approaches require 5200x more labels.

\section{DeviceScope: System Overview}
\label{sec:system}

\begin{figure}[tb]
    \centering
    \includegraphics[width=0.98\linewidth]{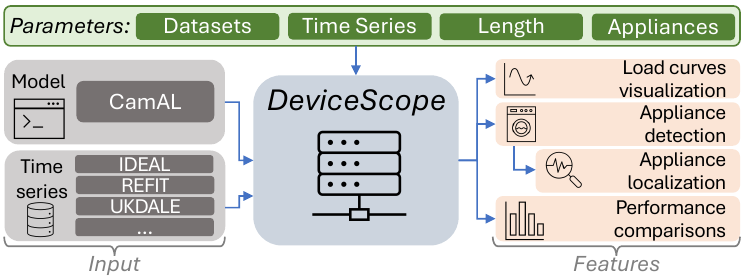}
    \vspace{-0.4cm}
    \caption{DeviceScope system inputs and features}
    \vspace{-0.5cm}
    \label{fig:system}
\end{figure}

In this section, we describe DeviceScope~\cite{DeviceScope_app}, the system developed to browse electrical consumption series, detect appliances, localize their probable activation time, and compare the performance of CamAL.
The GUI is a stand-alone web application developed using Python 3.10 and Streamlit. 
Figure~\ref{fig:system} illustrates the inputs and features of DeviceScope.

The (publicly available) datasets considered in our system are UKDALE, REFIT, and IDEAL~\cite{ukdale, refit, ideal} (though, users could upload other datasets, as well).
Each dataset comprises several houses monitored by sensors that record the total main and appliance-level power for a period of time (used only during evaluation).
We are interested in detecting five common appliances: the Kettle, Microwave, Dishwasher, Washing Machine, and Shower.
These appliances are selected because they have already been studied in the NILM literature and are suitable to be detected as they are not always ON. 

\begin{figure*}[tb]
    \centering
    \includegraphics[width=0.98\linewidth]{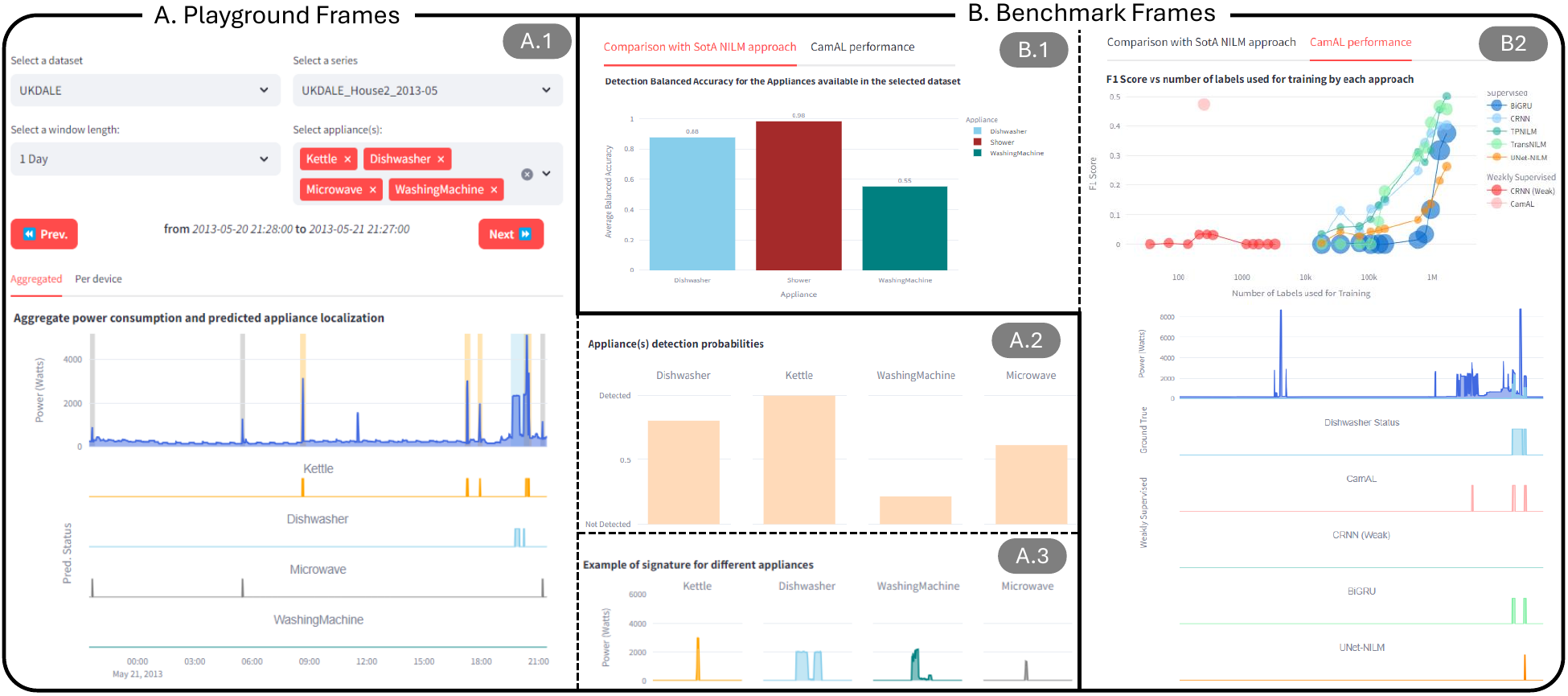}
    \vspace{-0.3cm}
    \caption{Illustration of the main frames of DeviceScope.}
    \vspace{-0.3cm}
    \label{fig:frames}
\end{figure*}

The user can select a dataset, a consumption time series, and a window length (6 hours, 12 hours, and 1 Day); once the series is loaded, the GUI displays the first sliding window following these parameters.
The user can then move to other adjacent windows using the {\bf Previous} and {\bf Next} buttons.
Then, the user can select an appliance to obtain the predicted status.
Finally, the user can browse the benchmark results.
We employ several measures to compare the models' performance regarding detection and localization, including Accuracy, Balanced Accuracy, Precision, Recall, and F1 Score.
First, the user can compare CamAL’s performance against 6 baselines based on the number of labels required for training.
Then, they can qualitatively assess its effectiveness on real data by evaluating appliance detection over a given period.


The GUI comprises two main frames, shown in Figure~\ref{fig:frames}.
The first frame (Figure~\ref{fig:frames} (A. Playground Frames)) is composed of 4 elements. 
(i) The first element (Figure~\ref{fig:frames} (A.1)) enables the user to display and navigate through the entire loaded consumption series data using the {\bf Prev.} and {\bf Next} buttons.
If a list of appliances is selected, the GUI will plot the predicted status for each appliance under the aggregate power consumption signal.
(ii) By selecting the {\bf Per device} view (Figure~\ref{fig:frames} (A.2)), the user can visualize the ground truth of individual appliance data and compare the predicted localization to real appliance consumption.
(iii) The {\bf Model detection probabilities} tab (Figure~\ref{fig:frames} (A.3)) enables the user to visualize the probabilities of detection of each selected appliance for the current window. 
The second frame (Figure~\ref{fig:frames} (B. Benchmark Frame)) is composed of 2 elements.  
(i) The first element (Figure~\ref{fig:frames} (B.1)) enables the user to browse the results of our method for each dataset interactively.
(ii) The second element (Figure~\ref{fig:frames} (B.2)) allows the user to compare CamAL’s performance with NILM methods based on the number of labels required for training. 
Additionally, it provides an interactive visualization of the predictions.


\section{Demonstration Scenarios}
\label{sec:scenario}

In this demo, we (i) propose a GUI to explore and visualize electrical consumption data to detect appliances in a given period, (ii) challenge the user to interactively localize appliance patterns and compare their estimation against the estimation obtained with CamAL (and also the ground-truth), and (iii) compare the performance and the number of labels needed to train CamAL compared to SotA NILM methods.

\noindent{\bf [Scenario 1: A blind guess]}:
This first scenario starts in frame 1 (Figure~\ref{fig:frames} (A. 1)), using the "Playground" page, where the user is invited to choose a dataset, load a time series, and select a window length.
They can then explore the electrical consumption data and try to identify which appliances have been used in the aggregate signal.  
For this scenario, the user has to guess without any external knowledge. 
The objective of this scenario is to demonstrate the challenging aspect of NILM without supervision.

\noindent{\bf [Scenario 2: A second guess with appliance patterns]}: For this second scenario, we will ask the user to guess again which appliances and when they have been used in the time series. 
The user is invited to open the "Playground" page (Figure~\ref{fig:frames} (A.1, A.2 and A.4)), load a time series, and select a window length. 
This time, we will ask the user to open the expander below the time series, depicting examples of appliance patterns.
The user can then select their appliance of interest in the corresponding select box to display the estimated localization appliance patterns by our system.
Last, the user can compare their guess, and the one predicted by our system, to the ground truth of individual appliance power consumption using the "Per device" tab.

\noindent{\bf [Scenario 3: Compare CamAL performance]} 
In this third scenario, the user is invited to select the "Benchmark" page (Figure~\ref{fig:frames} (B)) on the sidebar.
The user is invited to select a dataset, detection, and localization measure.
The users can click on the "Comparison with SotA NILM approach" tab to compare the performance of the different methods (6 baselines in total in addition to CamAL), as well as the number of labels needed to train CamAL compared to SotA NILM methods. 
In this same frame, the user can visualize each baseline prediction for each time series in our datasets. 
The latter allows the user to measure the strengths and limitations of the 7 approaches incorporated in our system and identify potential margins of improvement.  


\section{Conclusions}
\label{sec:concl}

We demonstrate DeviceScope, a system designed to help non-expert users better understand and easily analyze electricity consumption time series.
The system enables users to identify an appliance's use over a given period and estimate its use time using our novel weakly supervised CamAL approach.
DeviceScope enables electricity suppliers to easily identify which appliances the customer owns and their typical usage, thus allowing suppliers to propose personalized offers tailored to the customers' needs. 
It also helps customers save significantly by identifying over-consuming devices.

\section*{Acknowledgment}

Work supported by EDF R\&D, ANRT French program and EU Horizon projects AI4Europe (101070000), TwinODIS (101160009), ARMADA (101168951), DataGEMS (101188416) and RECITALS (101168490).

\bibliographystyle{IEEEtran}
\bibliography{sample}

\begin{thebibliography}{10}
\providecommand{\url}[1]{#1}
\csname url@samestyle\endcsname
\providecommand{\newblock}{\relax}
\providecommand{\bibinfo}[2]{#2}
\providecommand{\BIBentrySTDinterwordspacing}{\spaceskip=0pt\relax}
\providecommand{\BIBentryALTinterwordstretchfactor}{4}
\providecommand{\BIBentryALTinterwordspacing}{\spaceskip=\fontdimen2\font plus
\BIBentryALTinterwordstretchfactor\fontdimen3\font minus \fontdimen4\font\relax}
\providecommand{\BIBforeignlanguage}[2]{{%
\expandafter\ifx\csname l@#1\endcsname\relax
\typeout{** WARNING: IEEEtran.bst: No hyphenation pattern has been}%
\typeout{** loaded for the language `#1'. Using the pattern for}%
\typeout{** the default language instead.}%
\else
\language=\csname l@#1\endcsname
\fi
#2}}
\providecommand{\BIBdecl}{\relax}
\BIBdecl

\bibitem{DeviceScope_app}
\BIBentryALTinterwordspacing
A.~Petralia, P.~Boniol, P.~Charpentier, and T.~Palpanas. {DeviceScope}. [Online]. Available: \url{https://devicescope.streamlit.app/}
\BIBentrySTDinterwordspacing

\bibitem{eEnergy_ApplDetection}
A.~Petralia, P.~Charpentier, P.~Boniol, and T.~Palpanas, ``Appliance detection using very low-frequency smart meter time series,'' in \emph{e-Energy '23}, 2023.

\bibitem{themis_reviewnilm}
H.~Rafiq, P.~Manandhar, E.~Rodriguez-Ubinas, O.~{Ahmed Qureshi}, and T.~Palpanas, ``A review of current methods and challenges of advanced deep learning-based non-intrusive load monitoring (nilm) in residential context,'' \emph{Energy and Buildings}, 2024.

\bibitem{huquet_activation}
P.~Laviron, X.~Dai, B.~Huquet, and T.~Palpanas, ``Electricity demand activation extraction: From known to unknown signatures, using similarity search,'' in \emph{e-Energy '21}, 2021.

\bibitem{VLDB_TransApp}
A.~Petralia, P.~Charpentier, and T.~Palpanas, ``Adf \& transapp: A transformer-based framework for appliance detection using smart meter consumption series,'' \emph{PVLDB}, vol.~17, no.~3, 2023.

\bibitem{tsc_bakeoffredux_2024}
M.~Middlehurst, P.~Sch\"{a}fer, and A.~Bagnall, ``Bake off redux: a review and experimental evaluation of recent time series classification algorithms,'' \emph{Data Min. Knowl. Discov.}, vol.~38, 2024.

\bibitem{Wang2016TimeSC}
Z.~Wang, W.~Yan, and T.~Oates, ``Time series classification from scratch with deep neural networks: A strong baseline,'' \emph{IJCNN}, 2016.

\bibitem{dcam}
P.~Boniol, M.~Meftah, E.~Remy, and T.~Palpanas, ``Dcam: Dimension-wise class activation map for explaining multivariate data series classification,'' in \emph{SIGMOD '22}, 2022.

\bibitem{cam}
B.~Zhou, A.~Khosla, A.~Lapedriza, A.~Oliva, and A.~Torralba, ``Learning deep features for discriminative localization,'' in \emph{CVPR}, 2016.

\bibitem{gradcam}
R.~R. Selvaraju, M.~Cogswell, A.~Das, R.~Vedantam, D.~Parikh, and D.~Batra, ``Grad-cam: Visual explanations from deep networks via gradient-based localization,'' in \emph{2017 ICCV}, 2017.

\bibitem{ukdale}
J.~Kelly and W.~Knottenbelt, ``The uk-dale dataset, domestic appliance-level electricity demand and whole-house demand from five uk homes,'' \emph{Scientific Data}, vol.~2, 2015.

\bibitem{refit}
S.~Firth, T.~Kane, V.~Dimitriou, T.~Hassan, F.~Fouchal, M.~Coleman \emph{et~al.}, ``{REFIT Smart Home dataset},'' 2017.

\bibitem{ideal}
M.~Pullinger, J.~Kilgour, N.~Goddard, N.~Berliner, L.~Webb, M.~Dzikovska, H.~Lovell, J.~Mann, C.~Sutton, J.~Webb, and M.~Zhong, ``The ideal household energy dataset, electricity, gas, contextual sensor data and survey data for 255 uk homes,'' \emph{Scientific Data}, 2021.

\end{thebibliography}

\end{document}